\documentclass{article}
\usepackage{graphicx} % Required for inserting images
\usepackage{amsmath}
\usepackage{amssymb}
\usepackage{amsfonts}
\usepackage[round,authoryear]{natbib}
\usepackage[colorlinks=true,
            linkcolor=blue,
            citecolor=blue,
            urlcolor=blue]{hyperref}
\usepackage{cleveref}
\usepackage{subcaption}

\title{Understanding the Failure Modes of Transformers through the Lens of Graph Neural Networks}
\author{
  Hunjae Lee\\
  Department of Computer Science\\
  Southern Methodist University\\
  \texttt{hunjael@smu.edu}
}

\date{}

\begin{document}

\maketitle

\begin{abstract}
    Transformers and more specifically decoder-only transformers dominate modern LLM architectures. While they have shown to work exceptionally well, they are not without issues, resulting in surprising failure modes and predictably asymmetric performance degradation. This article is a study of many of these observed failure modes of transformers through the lens of graph neural network (GNN) theory. We first make the case that much of deep learning, including transformers, is about learnable information mixing and propagation. This makes the study of model failure modes a study of bottlenecks in information propagation. This naturally leads to GNN theory, where there is already a rich literature on information propagation bottlenecks and theoretical failure modes of models. We then make the case that many issues faced by GNNs are also experienced by transformers. In addition, we analyze how the causal nature of decoder-only transformers create interesting geometric properties in information propagation, resulting in predictable and potentially devastating failure modes. Finally, we observe that existing solutions in transformer research tend to be ad-hoc and driven by intuition rather than grounded theoretical motivation. As such, we unify many such solutions under a more theoretical perspective, providing insight into why they work, what problem they are actually solving, and how they can be further improved to target specific failure modes of transformers. Overall, this article is an attempt to bridge the gap between observed failure modes in transformers and a general lack of theoretical understanding of them in this space.
\end{abstract}

\section{Deep Learning through the Lens of Information Mixing and Propagation}
Much of modern deep learning can be understood as the study of learnable information mixing and propagation, a perspective that unifies seemingly disparate architectures under a common lens. Under this framework, many types of neural networks can be seen as parameterized functions that determine how information from different sources should be combined, weighted, and propagated to maximize performance on downstream tasks. 

Graph neural networks (GNNs) \citep{gilmer2017neural,hamilton2017inductive,xu2018powerful} exemplify this view most explicitly: node representations (containing node-level information) are iteratively updated by aggregating and transforming information from neighbors (i.e. neighbor nodes' representations), with learnable functions controlling how messages propagate along edges and mix at receiving nodes. A canonical high-level representation of a generic GNN can be written as follows:

\begin{align}
    h^{(l + 1)}_i = \phi(h^{l}_i, \sum^N_j A_{ij}\psi(h^{(l)}_i, h^{(l)}_j))\nonumber
\end{align}

where $\psi$ and $\phi$ denote aggregate and update functions, respectively. The learned representation of input data $x_i$ at layer $l$ is denoted $h^{(l)}_i$, with $h^{(0)}_i = x_i$. The aggregate function determines how to mix information from neighboring nodes with the current node while the update function determines how the current node's own representation will evolve given neighbor information and its current representation. The adjacency operator $A$ defines the graph topology (such as connectivity), determining the information mixing landscape.

Deep Sets \citep{zaheer2017deep} represents a more abstract form of information mixing neural networks where information from an unordered collection must be aggregated through symmetric operations, essentially mixing signals from multiple sources without dependence on their ordering. A generic invariant deep sets formulation can be written as:

\begin{align}
    h^{(l+1)}_i = \phi(\sum_i^{N}\rho(h^{(l)}_i))\nonumber
\end{align}

where both $\phi$ and $\rho$ denote universal approximators such as MLPs. This formulation of deep sets is strikingly similar to how the generic GNN formulation was derived above. Indeed, under the unified framework of information mixing, a GNN can be viewed as mixing with pre-defined paths (the graph structure dictates which nodes can mix with each other) whereas deep sets performs all-to-all mixing.

Transformers \citep{vaswani2017attention} take a step further by making the mixing weights themselves context-dependent: the attention mechanism learns which information sources (key-value pairs) to route and combine for each query position, with the propagation pattern dynamically adjusted by the input rather than fixed by graph structure (soft topology). Attention-based formulation of node (or token) update can be formulated as follows:

\begin{align}
    h^{(l+1)}_i = \phi(\sum^N_j (q^{(l)}_i \cdot k^{(l)}_j)v^{(l)}_j)\nonumber\\
    q^{(l)}_i = h^{(l)}_iW_Q, \; k^{(l)}_j = h^{(l)}_jW_K, \; v^{(l)}_j = h^{(l)}_jW_V\nonumber
\end{align}

Across all these architectures, their unifying objective is learning how to mix and propagate information flows to improve representations to ultimately solve downstream tasks. Under this framework, their specific inductive biases (graph topology, permutation invariance, full connectivity with learned routing, etc.) can be seen as determining what patterns of information mixing are prioritized or constrained. In fact, as has already been alluded, the distinction between these architectures is often arbitrary. Instead, there is a fluiditiy with which these architectures and their research domains shape and form one another, offering insights, interpretability, and improvements.

The quintissential example of this fluiditiy is the application of the attention mechansim into the GNN framework \citep{velivckovic2017graph}. As stated above in the discussion with transformers, attention mechanism can be seen as a soft topology, dynamically and learnably determining the strengh of propagation of different information sources. The application of this mechanism into the GNN framework has been incredibly successful, giving the model a way to learn to prioritize information from certain neighbors while dampening out irrelevant information from others. In addition, as we discuss later, attention mechanism is also provably more advantageous at certain information propagation tasks compared to uniform or symmetric information aggregators, making it's integration into the GNN space all the more significant. We provide a graph attention network (GAT) \citep{velivckovic2017graph,brody2021attentive}formulation below. 

\begin{align}
    h^{(l+1)}_i = \phi(\sum_j^{N}A_{ij}a_{ij} h^{(l)}_j), \quad a_{ij} = e(h^{(l)}_i, h^{(l)}_j), \; 0 < a_{ij} < 1\nonumber
\end{align}

Here, in addition to the adjacency operator $A$ defining the hard graph topology, there is now also $a_{ij}$ which learns the importance score for $h^{(l)}_j$ with respect to $h^{(l)}_i$, essentially controlling the rate of flow of information into $h^{(l)}_i$ from its neighbors $j \in \mathcal{N}_i$. In other words, the neighbors of $h_i$ are determined by the graph topology $A$, but how much information each neighbor contributes to $h_i$ is now dependent on the learned attention scores. 

Another example is the ideas in deep sets being used to influence the expressive power of GNNs. The graph isomorphism network (GIN) \citep{xu2018powerful} uses the permutation-invariant and sum-decomposition used in deep sets to create a GNN that is provably the most powerful among GNNs of its class. GIN is formulated as:

\begin{align}
    h^{(l+1)}_i = \phi((1 + \epsilon)h^{(l)}_i + \sum^N_jA_{ij}h^{(l)}_j), \quad A_{ii} = 0 \; \forall i \in \{N\}\nonumber
\end{align}

where the second term inside $\phi(\cdot)$ resembles the deep sets formulation and the self node update (diagonal entries $A_{ii}$ in the adjacency operator) is taken out of the sum decomposition and given a distinguishing mark in the form of $\epsilon$ to help the model distinguish the center node and its neighbors in all aggregation and updating steps. 

In addition to these examples, an increasingly favored approach in recent years has been to use GNNs to analyze, interpret, and even improve other architectures. Notably, studying the successes and more importantly, pitfalls of causal transformers that make up most modern LLMs through the lens of GNNs has been gaining in popularity. This is precisely what the bulk of this article focuses on. Before getting to causal transformers, however, a study of information mixing and propagation patterns in GNNs is necessary to understand the failure points of causal transformers, and LLMs at large.

\section{Information Propagation Issues in GNNs}
Broadly, there are three types of information propagation issues in GNNs: over-smoothing, over-squashing, and under-reaching. Over-smoothing \citep{li2018deeper,cai2020note,rusch2023survey} refers to a phenomenon whereby repeated application of information mixing (over many layers) causes all node features in the graph to become arbitrarily similar, losing expressive power and sharpness of information. Over-squashing \citep{alon2020bottleneck,topping2022understanding,di2023over}, on the other hand, refers to the compression of potentially exponentially growing information into fixed-sized representations, resulting in loss or distortion of certain information. Finally, under-reaching \citep{Barceló2020The} refers to situations where there are simply not enough layers for certain information to propagate to and mix with other information. The rest of this section expands upon these issues one-by-one.

\subsection{Over-smoothing in GNNs}
\begin{figure}[htbp]
  \centering
  \includegraphics[width=0.8\linewidth]{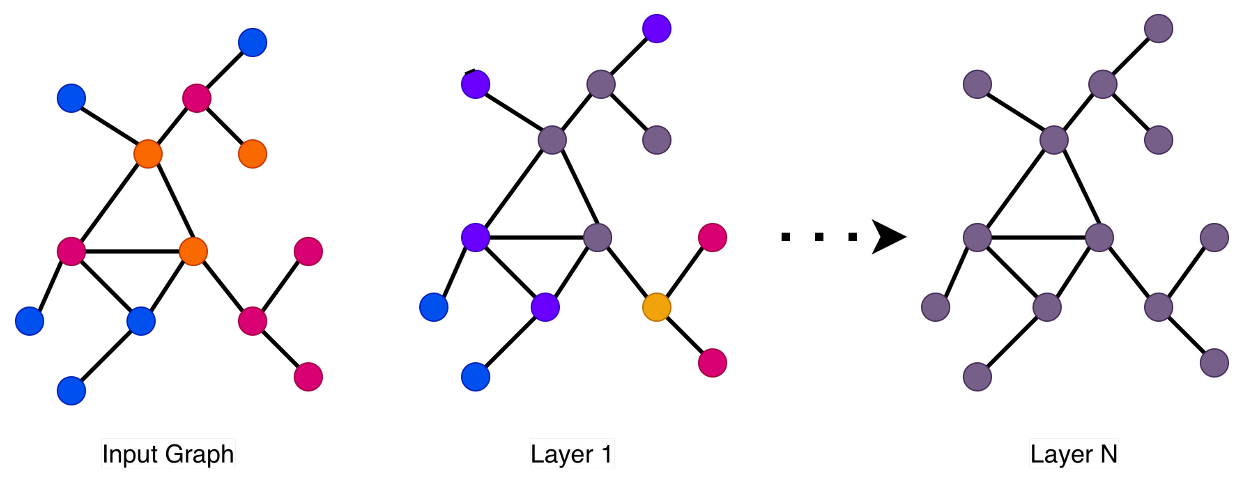}
  \caption{Over-smoothing in GNNs.}
  \label{fig:over-smoothing}
\end{figure}

An intuitive way to understand over-smoothing is through a color-mixing analogy. Imagine a color palette with arbitrary many dollops of paint where each dollop is a different color. As one can intuitively imagine, when they are mixed with a brush, they will initially form swirly patterns, retaining their individual colors but eventually converge to a single, possibly new, color. In the same way, over-smoothing in GNNs refers to a phenomenon where node representations in a graph converge to a common vector as a result of repeated information mixing operations as shown in \Cref{fig:over-smoothing}.

\subsubsection{Quanatifying and Characterizing Over-smsoothing}
Over-smoothing is often seen as a consequence of network depth (i.e. number of layers), whereby node representations become increasingly homegenous in deeper layers, ultimately causing nodes to become indistinguishable from each other. Formally, over-smoothing of node features $H^{(l)} = \{h^{(l)}_0, h^{(l)}_1,...,h^{(l)}_n\}$ in a graph can be characterized as:

\begin{align}
    lim_{l \rightarrow \inf} \delta(H^{(l)}) = c, \quad H^{(0)} = X_{\text{input}}\nonumber
\end{align}

where $\delta(\cdot)$ refers to some similarity measure between the node features and $c$ is some constant indicating the collapse of node embeddings. A commonly used measure of node feature similarity $\delta$ is Dirichlet Energy \citep{rusch2023survey}, defined as:

\begin{align}
    \delta_{DE} = \frac{1}{N}\sum_i\sum_{j \in \mathcal{N}_i}||H^{(l)}_i - H^{(l)}_j||^2_2\nonumber
\end{align}

which simply measures the $L_2$ distance between nodes and their neighbors, where lower values indicate more over-smoothing.

Intuitively, uniform and smooth information aggregation schemes such as mean-pooling or sum-pooling of neighbors suffers from over-smoothing sooner and more severely than sharp aggregation schemes like max-pooling. Attention mechanisms offer a good middle-ground as it is a weighted-sum, providing higher sharpness than mean/sum pooling while being smoother and retaining more information than max-pooling. It has been hypothesized by some that attention-based GNNs can prevent over-smoothing entirely due to its dynamic attention patterns controlling the flow of information in a learned manner. Intuitively, as attention can amplify and dampen different information sources, it can seemingly delay or prevent over-smoothing by simply learning information flow patterns that counteract progressive over-smoothing. However, more recent works have put an end to such hypothesis, proving definitively that graph attention networks (GATs) cannot prevent over-smoothing and in fact lose expressive power exponentially \citep{wu2023demystifying}. This is in part due to the continuous nature of softmax which prevents hard thresholding (i.e. softmax cannot return zero for any element no matter how long the sequence length). This means that over-smoothing eventually occurs in GATs, with it taking place faster if attention scores are relatively uniform (i.e. approaching mean aggregation) and slower if attention scores remain relatively sharp. 

\subsubsection{Mitigation Strategies for Over-smoothing}
Viewing over-smoothing as an inevitable consequence of information over-mixing has led GNN practioners to adopt practical solutions in model design to delay and mitigate its negative effects. One such example is simply reducing the number of layers in GNNs. Indeed, GNNs are often shallow, with the number of layers rarely exceeding 6. In fact, the graph topology is often a strong enough inductive bias that GNNs with only 2 or 3 layers are performant enough across many tasks. Another solution to mitigate over-smoothing is using residual connections. Residual connections have often been used in an ad-hoc fashion such as in transformers \citep{vaswani2017attention} without clear interpretive reasons. However, under the over-smoothing framework, residual connections can be seen as slowing down the effects of information over-mixing by letting each node (or token) be reinforced with its own representation after each round of information-mixing, allowing the retention of its own identity for longer. This represents the first of many instances covered in this article where analyzing information propagation issues of GNNs offer interpretable and insightful perspective to transformers and LLMs. 

\subsection{Over-squashing in GNNs}
\begin{figure}[tbp]
  \centering
  \includegraphics[width=0.8\linewidth]{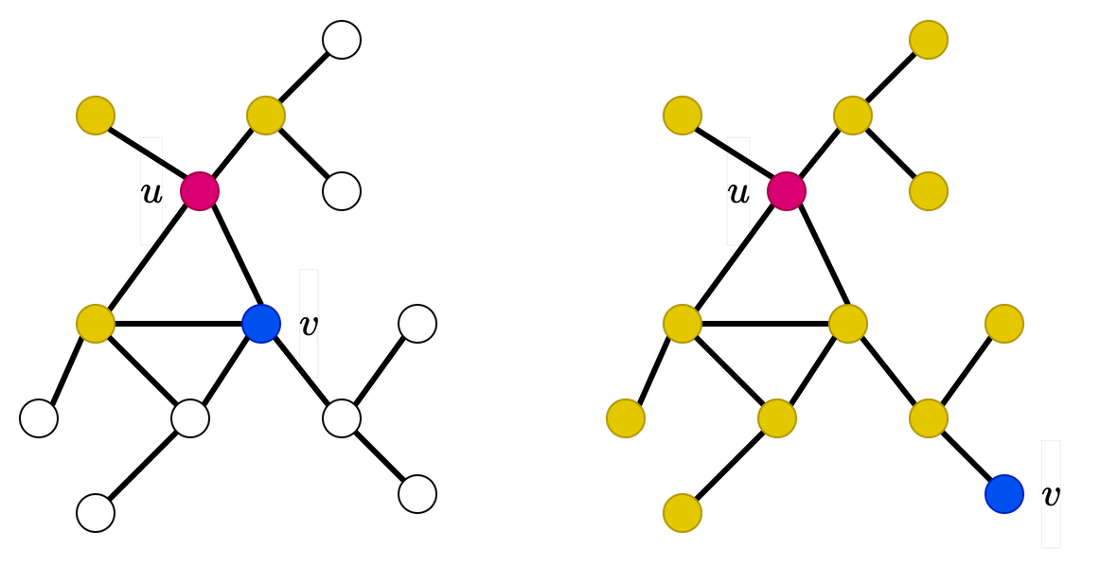}
  \caption{Over-squashing in GNNs.}
  \label{fig:over_squashing}
\end{figure}

Over-squashing refers to representation bottlenecks that occur when too much information is compressed into fixed-sized node representations, resulting in distortion or even loss of information. This was initially observed in RNNs, where the sequential nature of information propagation resulted in final representation being over-compressed or "squashed out" by the linearly growing receptive field of previous information. In GNNs, this was first observed by \citep{alon2020bottleneck} where it was noted that this effect is strictly more harmful in GNNs, whose receptive field grows exponentially, not linearly. In fact, they observe that for tasks where sufficiently long-range information propagation is necessary, GNNs can not even fit the training data even with sufficient number of layers. Receptive field of a node in this context just means the amount of information (number of nodes) that must be mixed and propagated into the receiving node. For example, in one layer, each node in a regular graph with degree of $k$ has a receptive field of $k$ (its neighbors). Over 2 layers, each node's receptive field grows to $k^2$ nodes, and in general over $l$ layers, each node has a receptive field of $k^l$ nodes. Consider the example in \Cref{fig:over_squashing}. On the left, the two nodes $u$ and $v$ are only one hop apart, making their receptive field only 3 other nodes. In contrast, on the right where the two nodes are three hops apart, the receptive field becomes the entire graph. As such, distant nodes are at a disadvantage and risk getting squashed out of the information mixing landscape.

\subsubsection{Bounding Over-squashing with the Jacobian}
Over-squashing is not a product of model depth like over-smoothing but rather a product of graph topology itself. In particular, long-range information propagation tends to suffer as information from distant nodes get squashed out by the exponentially growing receptive field. Intuitively, long-range information propagation causes over-compression of information and ultimately loss of desired signal at a long-enough propagation distance. However, over-squashing is not defined by long-range information propagation alone, and can actually occur between nodes that are relatively close to each other in topological distance. This is precisely what makes the study of over-squashing open-ended and difficult to characterize in a clean and reductive manner.

\begin{figure}[bp]
  \centering
  \includegraphics[width=0.8\linewidth]{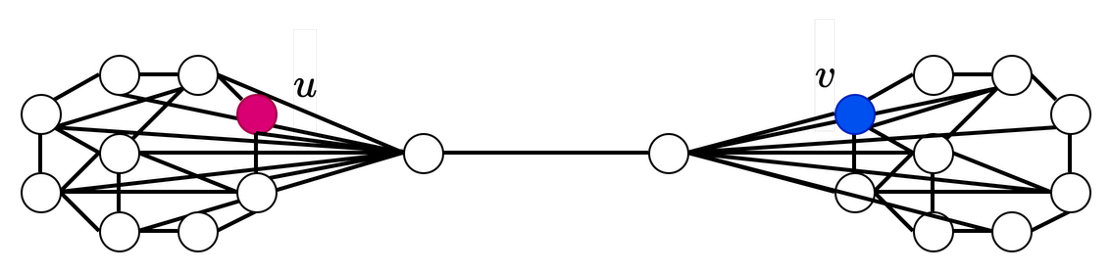}
  \caption{Barbell graph.}
  \label{fig:barbell_graph}
\end{figure}

One instance where short-range information propagation can suffer over-squashing is shown in \citep{topping2022understanding}, where it is demonstrated that negatively curved edges can cause a certain type of over-squashing driven by topological bottlenecks rather than distance. For an intuition on negatively curved edges and how they affect over-squashing, consider a barbell graph in \Cref{fig:barbell_graph} where two dense subgraphs are connected by few edges called bridges. While these bridges enable short-range information propagation, the fact that the sum total of each subgraph's dense information have to go through relatively few edges mean that information over-squashing invariably occurs at these bridges. Therefore, even at short distances, the receptive field can blow up and cause over-squashing. 

It is clear that over-squashing is a difficult idea to characterize since different graph topologies can present radically different information propagation patterns. This is addressed in \citep{topping2022understanding} where a ground-truth metric is provided to classify node-level over-squashing. They use the Jacobian of the node features to quantify the amount of over-squashing being experienced at a given node as follows:

\begin{align}
    |\frac{\partial h^{(r+1)}_i}{x_s}| \le (\alpha\beta)^{r+1}(A^{r+1})_{is}\nonumber
\end{align}

where node $i$ and $s$ are exactly $r+1$ distance apart, $||\nabla \phi|| \le \alpha,\; ||\nabla \psi|| \le \beta$, and $A^{m}$ is the adjacency operator at the $m^{\text{th}}$ power. The Jacobian measures the sensitivity of a given node $h_i$ at layer $l=(r+1)$ to perturbations in some node $x_s$ at the input layer $l=0$. Intuitively, the more node $i$ remain sensitive to node $s$, the less over-squashing is experienced. On the other hand, if changes to node $s$ has little to no effect on the representation of node $i$, it is likely experiencing over-squashing. The upper bound of the Jacobian can be interpreted as a node's capacity for over-squashing being affected by both model design ($\alpha$ and $\beta$) as well as the graph topology $A$.

Extending from the bounds introduced in \citep{topping2022understanding} which required nodes to be exactly $r$ hops apart, \citet{black2023understanding} introduces a Jacobian bound with respect to any two nodes in the graph at arbitrary distance, as follows:

\begin{align}
    |\frac{\partial h^{(r)}_u}{\partial x_v}| \le (2\alpha\beta)^r\sum_{l=0}^{r}(A^l)uv\nonumber
\end{align}

These bounds from \citep{black2023understanding} can be further generalized to quantify over-squashing bounds between any two nodes $h_u$ and $h_v$ at arbitrary layers $r + L$ and $r$ with $L > 0$,, as follows:

\begin{align}
    |\frac{\partial h^{(r+L)}_u}{\partial h^{(r)}_v}| \le (2\alpha\beta)^L\sum_{l=r}^{r+L}(A^l)_{uv}\nonumber
\end{align}

This provides layer-aware profiling on information flow across the network and shows that over-squashing is not just a global quantity over all layers but can occur within different layers as well. The layer-agnostic bound is derived by the author of this article and is not part of the formulation in \citep{black2023understanding}.

Concurrent to \citep{black2023understanding}, a more model-specific bound is introduced in \citep{di2023over}, accounting also for the effect of model dimensions on over-squashing. This is expressed as follows:

\begin{align}
    |\frac{\partial h^{(r+1)}}{x_s}| \le (c_{\sigma}wd)^{r+1}(c_{\alpha}I + c_{\beta}A)\nonumber
\end{align}

where $c_{\sigma}$ is the bound on the activation function, $c_{\alpha}$ and $c_{\beta}$ represent the gradient bounds for update and aggregation functions, $w$ denotes maximum magnitude of the weight matrices and finally $d$ represents the model dimensions, or the width of the model. This bound implies that models with larger width (model dimensions) can mitigate over-squashing. However, as this work points out, increasing width to mitigate over-squashing is a tricky balance as it tends to increase model over-fitting.

The Jacobian-based sensitivity measure is typically considered the most faithful and definition-aligned measure of over-squashing and offers theoretical avenues to bound and analyze over-squashing effects. However, the Jacobian bounds are almost never used directly in practice because they are computationally and statistically impractical at scale. Instead, to practically quantify over-squashing, GNN practitioners often fall back on tractable metrics such as effective resistance, commute time, or spectral gap of the graph Laplacian. These proxy quantities provide practical and tractable metrics for over-squashing, which are crucial in mitigation strategies for over-squashing such as graph rewiring methods. 

\subsubsection{Tractable Metrics for Over-squashing and Graph Rewiring}
Tractable proxies for Jacobian-based over-squashing bounds are typically graph-theoretic quantities that measure how easily information can flow across bottlenecks. As a result, these quantities are widely used to drive graph rewiring. Graph rewiring \citep{arnaiz2022diffwire,karhadkar2023fosr} is a process whereby the graph topology is modified during training or as a pre-processing step typically to reduce bottlenecks and improve information flow. Rewired graphs are often used in conjunction with the input graph topology to maintain some semblance of the input topological inductive bias. This section covers several of these proxy quantities such as curvature, effective resistance, commute time, and the spectral gap.

Curvature-based metrics (typically Ricci or Foreman curvature bounds) assign local scores to edges, where highly negative curvatures denote narrow cuts and thus likely representing information bottlenecks. In rewiring strategies, these curvature-based metrics drive the model to iteratively add edges to reduce negative curvatures to alleviate topological bottlenecks and thereby ultimately reducing over-squashing. 

The spectral gap $\lambda_1$ is the smalllest non-zero eigenvalue of the graph Laplacian and is another measure of topological bottlenecks in a graph. In actuality, it is yet another quantity, called the Cheeger constant $h_G$ \citep{Chung1996SpectralGT} which measures how hard it is to separate a graph into two disjoint sets. This is also called the cut of a graph and has useful properties relating to flow networks. The Cheeger constant, however, is impractical to calculate. Thankfully, the Cheeger constant bounds the first non-zero eigenvalue of the normalized graph Laplacian as follows:

\begin{align}
    \frac{h_G^2}{2} \le \lambda_1 \le 2h_G\nonumber
\end{align}

This is known as the Cheeger inequality and $\lambda_1$ is referred to as the spectral gap of the graph. Since small $h_G$ implies more topological bottlenecks, increasing the spectral gap is often used as a heuristic to decrease over-squashing. In this way, the spectral gap is sort of a proxy within a proxy for measuring over-squashing.

Clearly, there is a relationship between curvature and the spectral gap. While both quantities measure local topological bottlenecks, curvature metrics may provide more edge-wise information since they assign a score to each edge, while the spectral gap can be seen as a graph-level summary of local topological bottlenecks.

\begin{figure}[tbp]
  \centering
  \includegraphics[width=0.8\linewidth]{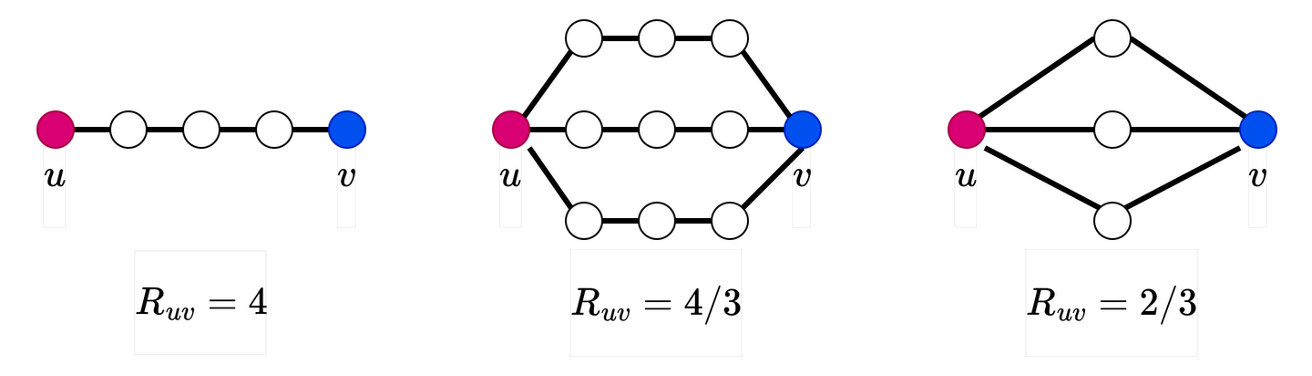}
  \caption{Effective resistance profiling on different graphs.}
  \label{fig:effective_resistance}
\end{figure}

In contrast, effective resistance $R_{uv}$ and its random-walk counterpart, commute time $C_{uv}$, provide global pairwise measures of topological bottlenecks induced by the graph geometry. Effective resistance is an idea borrowed from electrical circuits, where instead of measuring electrical flow, effective resistance for graphs measures flow of information. Large $R_{uv}$ implies that information from nodes $u$ and $v$ must traverse many shared edges to reach each other with relatively small number of paths to reach each other. As such, large effective resistance implies bottlenecks. To provide an intuition on how effective resistance classifies different graphs, we provide some examples in \Cref{fig:effective_resistance}. From left to right, effective resistance decreases with more paths and less shared edges. Commute time $C_{uv}$, on the other hand, measures the expected number of steps for information from node $u$ to reach $v$ and return back to $u$. Because both effective resistance and commute time account for all paths for each node-pair, they are often considered more informative than curvature metrics, which are local by nature. These two quantities are tightly linked by a mathematical relationship $C_{uv} = 2mR_{uv}$ in an unweighted graph with $m$ edges. Large values in both effective resistance and commute time are provably associated with small Jacobian entries and therefore, more over-squashing. Rewiring approaches for effective resistance and commute time often add or delete edges iteratively to reduce such quantities. 

There are also graph rewiring approaches that do not use any bottleneck or over-squashing measures and instead rely on graph-theoretic principles to design computational graphs as a pre-processing step \citep{deac2022expander,wilson2024cayley}. These approaches often use expander graphs, which are graphs with high connectivity despite having sparse edges, allowing for rapid mixing of information with relatively few edges. The precise graph-theoretical properties that enable expander graphs are beyond the scope of this article. Expander graphs are often used in conjunction with the original graph during training and inference to maintain the inductive bias of the input topology. Other than expander graphs, incorporating a fully-connected layer was found to improve over-squashing considerably \citep{alon2020bottleneck}. Notably, replacing the graph topology entirely and only using fully-connected graph for all layers (effectively turning it into a transformer) was found to result in poor performance, indicating that input graph topology is still important for downstream tasks even if it carries information propagation issues. Instead, using a fully connected graph only at the last layer resulted in the best performance, alleviating over-squashing while giving the model a chance to use the input graph topology as an inductive bias in early layers. 

\subsection{Under-reaching in GNNs}
\begin{figure}[tbp]
  \centering
  \includegraphics[width=0.8\linewidth]{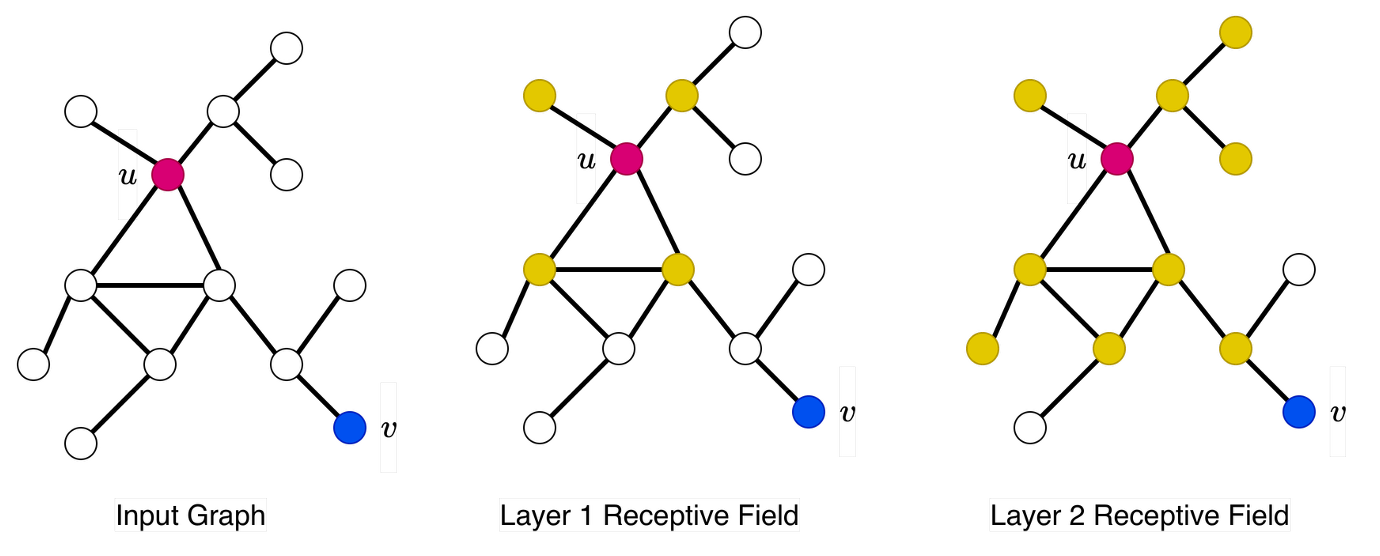}
  \caption{Under-reaching in GNNs.}
  \label{fig:under_reaching}
\end{figure}

Under-reaching in GNNs refers to when the depth of the message-passing architecture is insufficient to cover the problem radius which is the maximum graph distance over which task-relevant dependencies must be propagated. This results in many required interactions simply never being “seen” by the model. For an $L$-layer GNN, each node representation $h^{(L)}_i$ is a function only of nodes in its $L$-hop neighborhood, so if the task depends on signals at distance $r > L$, the corresponding Jacobian entries $\frac{\partial h^{(L)}_i}{\partial x_s}$ vanish entirely for all pairs with $dist(u,v) > L$. For example, \Cref{fig:under_reaching} shows a graph where a 2-layer GNN cannot mix nodes $u$ and $v$. This happens not because information was squashed (over-squashing) or over-mixed to convergence (over-smoothing), but because it never entered the computation graph. This makes under-reaching an architectural limitation. Even with infinite width (model dimensions) and perfect optimization, no choice of parameters can recover dependencies beyond $L$ hops, and increasing parameter count or improving bottlenecks cannot fix a mismatch between depth and problem radius. In practice, under-reaching manifests when shallow GNNs plateau on tasks requiring long-range information mixing. 

Unlike over-squashing and over-smoothing, under-reaching is a much more binary problem in that it is completely rectified when the model has enough layers for the given task. On the other hand, if there are not enough layers for tasks where long-range information mixing is crucial, the performance may not degrade gracefully but simply collapse. In general, under-reaching is not a big issue in modern GNNs. In transformers with all-to-all connection where each token is connected to every other token, under-reaching never occurs. While under-reaching is not an issue for transformers, it is still included in this article as it highlights how information mixing and propagation are inextricably tied to model design and the topological profile of the problem. With transformers being so performant out of the box, transformer researchers may have never even considered a problem like under-reaching. However, through the lens of GNNs, many advantages of transformers are given an explanation which may pave the road for even better solutions in the future. 

When an understanding is achieved on why something works as well as it does, it can fuel further improvements. More importantly, as discussed in-depth in the remainder of this article, analyzing transformers through the lens of GNNs can uncover many failure points of transformers, explain counter-intuitive behaviors of transformers under the hood, and shed light on how they can be improved.

\section{Information Propagation Issues in Transformers}
As alluded in the first section, transformers and GNNs are two sides of the same coin. In fact, a transformer layer can be written exactly as a message-passing update on a fully connected graph of tokens, with self-attention implementing a learnable, content-dependent neighborhood aggregation \citep{joshi2025transformers}. In other words, a standard transformer is a graph attention network (GAT) with a fully connected graph. This fact alone gives insight into many aspects of transformer design such as why positional encodings have been found indispensible in transformers for LLMs. Because GNNs are permutation-invariant by design, positional encodings in transformers can be seen as an attempt to break this permutation-invariance and to preserve the sequential nature of tokens. More importantly, motivating transformes as an instance of GNNs naturally raises the question of if and how the information propagation issues of GNNs concern transformers for LLMs. Intuitively, under-reaching is never an issue in transformers since all connections are exactly 1-hop away, meaning any transformer with one or more layers suffices in eliminating under-reaching entirely. However, as is discussed in-depth in this section, both over-smoothing and over-squashing affect transformers for LLMs. In addition, the use of causal masking in decoder-only transformers which make up most of modern LLMs create interesting geometric properties and cause a predictably asymmetric flow of information. By analyzing transformers under the lens of GNN theory, inadvertent observed side-effects of transformers such as attention sinks, last-token representation collapse, observed U-shaped performance in retrieval tasks, and more can be unified under a geometric interpretation of the causal graph.

\subsection{Over-smoothing in Transformers}
Over-smoothing effects have been emprically observed and studied in transformers independently to GNNs and is known by another name: rank collapse. GNN practitioners with their knowledge of over-smoothing have since provided a lot of theoretical insight into this phenomenon in transformers. In fact, recent works have shown that with pure self-attention, as long as there is a token which all other tokens in the sequence can directly or indirectly attend to over a fixed number of layers, exponential rank collapse of tokens to a common vector is guaranteed \citep{wu2024role}. Akin to over-smoothing in GNNs, rank collapse in transformers can be seen through the lens of information over-mixing. As a result, sparse or local attention can provably slow down the collapse rate, but not prevent it entirely. In addition, residual connections and layer normalizations can also aid in slowing down the collapse rate. 

Rank collapse or over-smoothing in transformers is also exacerbated by the very nature of attention scores and the mechanics of the softmax function. As discussed above in over-smoothing analysis for GATs, softmax assigns a non-zero attention score to every element in the sequence no matter how long the sequence length is. This means that the attention map can be seen as a mixing matrix whereby its repeated application can only drive representations closer to each other in value but never drive them apart. In other words, token representations are at their most distinct in the input layer, and are made more and more homogenous with increased layers. This is not necessarily problematic in isolation since a degree of homogeneity is desirable for stable embeddings and for tokens to share information with each other. Rather, this only becomes problematic when driven to sufficient loss of discriminative power or total convergence of token embeddings.

\subsection{Geometry of the Causal Graph}
Decoder-only transformers make up most modern LLMs. Unlike a regular fully-connected transformer, a decoder-only transformer is defined by its causal masking. Causal mask is used to ensure that next-token prediction setup is not compromised by tokens having access to tokens in later positions during training. Causal mask $M \in \mathbb{R}^{N\times N}$ is an upper-triangular matrix filled with $-\inf$ applied to pre-softmax logits to ensure that softmax returns 0 for maksed out positions in each row and is formulated as follows:

\begin{align}
    \text{Attention}(X) = softmax(\frac{XW_QXW_K^T}{\sqrt{d_k}} + M)XW_V\nonumber
\end{align}

On the surface, taking the full transformer and applying the causal mask seems to be a relatively inconsequential, if not clever solution. After all, what drastic change would occur just because half of the outputs of the transformer get masked out? As a result, for a long time there was not much research being done to analyze precisely how the causal mask changes the geometry of information propagation in transformers. In fact, studying information propagation issues in general had been a relatively niche research area confined mainly to GNN researchers. 

\begin{figure}[t]
  \centering
  \begin{subfigure}[t]{0.45\textwidth}
    \centering
    \includegraphics[width=\textwidth]{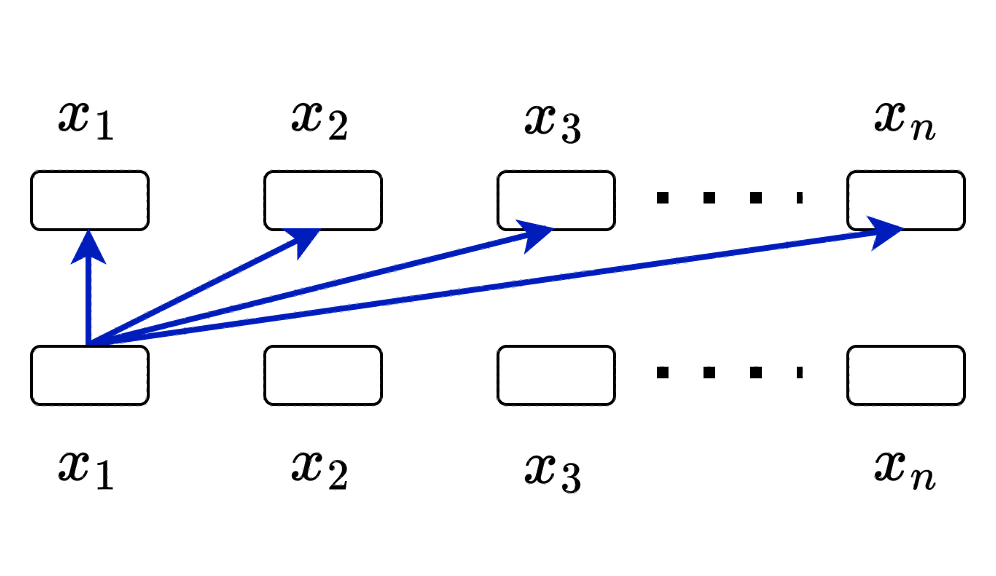}
    \caption{}
    \label{fig:first_token}
  \end{subfigure}
  \hfill
  \begin{subfigure}[t]{0.45\textwidth}
    \centering
    \includegraphics[width=\textwidth]{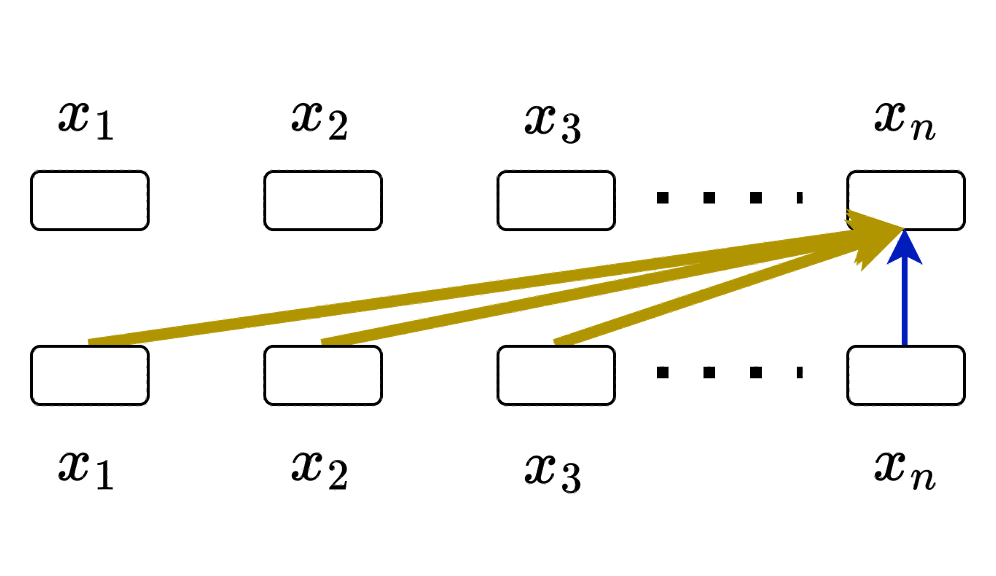}
    \caption{}
    \label{fig:last_token}
  \end{subfigure}
  \caption{Causal graph of the (a) first-token and (b) last token.}
  \label{fig:tokens}
\end{figure}

However, when viewing causal masking as a graph, the issues of information propagation in decoder-only transformers become clear. As expected, it was GNN practitioners who began noticing that causal masking turns transformers from a fully-connected graph into a directed, bipartite one with an unmistakable position bias created as a result. Position bias occurs when a model is biased toward focusing on certain regions of the input. In decoder-only transformers, tokens at every position can only attend to tokens at current and previous positions in the sequence. In other words, with token indices denoting their position in the sequence where $X = \{x_1,x_2,...,x_n\}$, token at position $i$ can only attend to tokens at positions $j \le i$ making its receptive field $\{x_1,x_2,...,x_i\}$. What this means is that tokens toward the beginning of the sequence have many opportunities to mix and refine their representations while tokens toward the end of the sequence have very few opportunities to do the same. This is sometimes called the runway problem whereby tokens earlier in the sequence have more "runway" to refine their representations while later tokens experience runways that are increasingly shorter. As shown in \Cref{fig:tokens}, in the extreme case, the last token in the sequence $x_n$ has only one chance to refine its representation and has the largest receptive field containing every token in the sequence when updating its own representation. In contrast, the first token in the sequence $x_1$ is attended to by every token in the sequence and thus influences every token's representation.

Therefore, causal masking inherently biases attention toward earlier positions \citep{wu2025emergence}. Tokens in deeper layers attend to increasingly more contextualized representations of earlier tokens which amplify their influence in the information-mixing landscape. In fact, the first token of the sequence can be interpreted as a center node, as pointed out by \citep{wu2025emergence}. A center node in graph theory is a node from which every node in the directed graph is reachable. In information-mixing terms, center nodes carry an inherent topological advantage that when taken to the limit, will dominate the overall representation landscape. As a result, information propagation patterns in decoder-only transformers exhibits a predictable asymmetry, favoring earlier positions. 

While this is an inevitable consequence of causal masking, it is still problematic for a number of reasons. First, many studies show that this results in attention patterns where early positions get very high attention scores even if they are semantically irrelevant \citep{xiao2023efficient,gu2024attention}. This has been named as the attention sink phenomenon \citep{xiao2023efficient}. Second, and perhaps more intuitive, sequences of words or tokens do not inherently have position bias. When writing sentences, earlier words do not carry more meaning than words later in that sequence. As such, causal graphs create a fundamental mismatch between the actual value and importance of tokens with respect to the sequence and how the model views token importance. Finally, the geometry of the causal graph creates situations where last token in the sequence can experience severe over-squashing, also called representation collapse, as illustrated in \Cref{fig:last_token}. This is particularly problematic since in next-token prediction schemes, the last token in the sequence is solely responsible for predicting the next token. In other words, in decoder-only transformers, the token that is provably experiencing the most over-squashing is also the one solely responsible for generating the next token. The rest of this section discusses these issues created by the geometry of causal masking in-depth.

\subsection{Attention Sinks and the Uniqueness of the First Token}
In a causal transformer, attention sinks and the uniqueness of the first token both arise from the strong position bias induced by the geometry of the causal graph. For the first token, a lower-triangular attention pattern means that its represention is never updated by other tokens through attention and only evolves vertically via residual connections and the MLP stack, while all later tokens remain free to attend back to it. This means that the first token in each sequence is never actually updated via the attention mechanism. 

This structural phenomenon, a direct result of causal masking, makes the first token a natural global anchor: it has maximal out-degree (every later position can attend to it) but trivial in-degree (it never sees any other token for its own representation update), so during training the model is incentivized to route surplus or stabilizing information through this token, turning it into an attention sink that persistently absorbs a large fraction of attention scores independent of local semantics. This was initially discovered by \citep{xiao2023efficient}, noting attention sinks as the emergence of strong attention scores towards initial tokens as a sink even if they are not semantically important. They note that adding a placeholder token as a dedicated attention sink during training can improve performance. In addition, \citep{xiao2023efficient} attribute the reason for the emergence of attention sinks to the softmax function, which requires attention scores to sum up to one in each row of the lower-triangular attention map. Even when a given token does not align well to the previous tokens, the model still needs to allocate unneeded attention scores somewhere so it sums up to one, making the always visible initial tokens ideal pathways to "soak up" unneeded attention. Other works confirm this phenomenon, showing that early tokens learn to develop representations that attract high levels of attention scores while keeping their value norms small so as to not influence the representation landscape with their high attention scores \citep{gu2024attention}. 

While attention sinks are certainly an unintended consequence of the position bias created by causal masking, it is not clear if attention sinks are necessarily problematic. For instance, \citep{gu2024attention} notes that forcibly trying to remove the attention sink phenomemon leads to performance degradation. Furthermore, \citep{barbero2025llms} notes that attention sinks may be providing a mechanism for LLMs to avoid over-mixing by soaking up large portions of attention scores, slowing down information mixing as a result. This suggests that attention sinks may actually have positive consequences, potentially slowing down over-smoothing in causal transformers and resulting in more stable token embeddings. The study of attention sinks is very much a recent development and as a result, lacks definitive answers. The literature is in congruence that having dedicated sink tokens injected into the sequence seem to be a practical solution, achieving the benefits of attention sinks while not sacrificing the actual first tokens in the sequence to become attention dumps. 

\subsection{Over-squashing and the Runway Problem in Transformers}
\begin{figure}[tbp]
  \centering
  \includegraphics[width=0.8\linewidth]{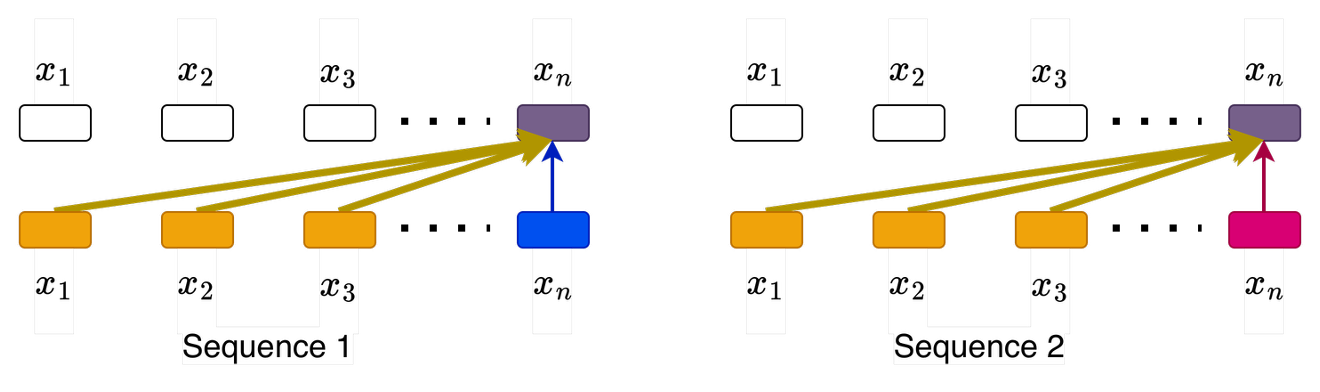}
  \caption{Last token representation collapse.}
  \label{fig:last_token_representation_collapse}
\end{figure}

The position bias of causal transformers create two related problems for tokens in later positions: over-squashing and the runway problem. Over-squashing, also called representation collapse in transformer literature, occurs in tokens toward the end of the sequence as the receptive field grows as a consequence of causal masking. In fact, it has been shown that last-token representations of different sequences can become arbitrarily close, losing information sharpness \citep{barbero2024transformers}. This occurs when two sequences have identical tokens throughout the sequence except for the last position, as shown in \Cref{fig:last_token_representation_collapse}. The distinct last token representation gets increasingly drowned out by the growing receptive field as the sequence length grows, until it loses all sharpness and the two distinct last token representations from both sequences become arbitrarily close to each other.

As mentioned earlier in this section, this is particularly problematic for next-token generation tasks, which make up the overwhelming majority of LLMs. In next-token generation schemes, the last token in the sequence is solely responsible for generating the next token. This means that the token experiencing the most over-squashing is the one solely being used for the actual task at hand. For multiple-token generation (as is the case in chat-based LLMs for example), it is simply an iterative process of the last token in the sequence generating the next token, carrying on the representational defects of the last token iteratively for each generation. 

The runway problem refers to the fact that tokens earlier in the sequence have more opportunities to preserve and refine their representations while tokens later in the sequence have increasingly shorter runways. This is pointed out by \citep{barbero2024transformers} as well as \citep{wu2025emergence}, who further mentions that tokens in deeper layers attend to increasingly more refined representations of earlier tokens, thereby amplifying the influence of initial positions. With sufficiently deep layers, the initial positions' information mixing dominate all representations. 

Intuitively, the runway problem of causal transformers may have connections to effective resistance in \citep{black2023understanding} since effective resistance is a measure of well-connectedness of nodes concerning how many paths exist as well as how short such paths are. Under this lens, tokens coming earlier in the sequence may be seen as having lower effective resistance than tokens later in the sequence as they have more paths to reach the end of the sequence. This may offer yet another bridge between GNN theory and over-squashing research in transformers, advancing the conversation in a meaningful, theoretically grounded way. However, the effective resistance formulation in \citep{black2023understanding} is confined to non-bipartite graphs, excluding causal graphs as a consequence. In addition, the linear mathematical relationship between effective resistance and commute time further complicates the issue since commute time is not defined for causal graphs. 

\subsection{Interpreting Ad-hoc Solutions with Theory}
With theoretical knowledge into both GNNs and transformers, existing solutions in causal transformer research can be better contextualized. These solutions are often presented with little to no theoretical motivation and are typically motivated purely by intuition. By contextualizing them theoretically, their successes can be better understood, and may pave the way for better solutions in the future now that there is a clear understanding of what problem is actually being solved.

First is the introduction of pause tokens by \citep{goyal2024think}. They inject learnable pause tokens (fresh computation space) to allow the model to process extra computation before committing to an answer (i.e. next token prediction). Instead of manipulating $k$ tokens to determine the $(k+1)$th token representation, they essentially increase the context with pure learnable tokens. With 10 pause tokens for example, $(k+1)$th token representation is determined by $k+10$ tokens. They justify this simply as giving the model extra time to think without a convincing theoretical justification or motivation. However, with more knowledge into GNN and transformer theory, pause tokens can be understood as counteracting both over-smoothing and the runway problem. Similar to attention sinks, pause tokens may slow down information over-mixing by providing attention dumps in the form of blank, learnable pause tokens which allow actual tokens to remain sharper for longer. This may end up slowing down the effects of over-smoothing (or rank collapse) and thereby improve performance. Perhaps more importantly, pause tokens can be seen as elongating the runway especially for tokens later in the sequence. By appending pause tokens at the end, the actual last tokens in the sequence get more runway to refine and preserve their information, instead of getting squahsed out in one fell swoop. The authors observe that appending pause tokens at the end result in better performance than pre-pending them to the beginning of the sequence, which is precisely the expected result when coming from the perspective of the runway problem. Overall, pre-pending pause tokens in the beginning of the sequence may improve over-smoothing in an attention sink like manner while appending pause tokens at the end may improve the runway problem and reduce over-squashing. As such, under more theoretical motivation, pause tokens may serve a very different purpose depending on whether they are put in the beginning or the end of the sequence and thus is not a straightforward design choice to choose one or the other. With this theoretical contextualization, a potentially better solution can be formulated in the form of a general padding around both the beginning and end of each sequence with learnable, semantically blank tokens.

Next, the differential transformer can be better positioned with theoretical justification. The differential transformer \citep{ye2025differential} is a method where an additional, second attention map is learned and used to subtract from the main attention map as shown below.

\begin{align}
    \text{DifferentialAttention}(X) = (softmax(\frac{Q_1K_1^T}{\sqrt{d}}) - \lambda softmax(\frac{Q_2K_2^T}{\sqrt{d}}))V\nonumber
\end{align}

where $[Q_1;Q_2] = XW_Q, \; [K_1;K_2] = XW_K, \; V = XW_V$ and $\lambda$ controls the strength of the subtractive map.

The technique is motivated as a "de-noising" process, where the second attention map learns to act like a noise-cancelling signal, cancelling out the noise of the main attention map and thereby keeping relevant signals sharper. While this is certainly a plausible explanation, in theoretical terms, this may yet again be explained as a technique to slow down information over-mixing and mitigate model over-smoothing. Differential transformer is notably better in retrieval tasks as well as general model performance compared to standard transformers. In addition, differential transformer may be better positioned to specifically reduce over-squashing if the subtractive attention map becomes a function of the previous rows instead of solely the current row. As it stands, for each row, what is being subtracted is a function of the current row. This means that the de-noising effect only applies to noise at the current row. However, to tackle accumulated noise as a result of position bias, the subtractive map should be made a function of the prior information mixing landscape. 

\section{Conclusion}
Despite their potentially devastating implications, decoder-only transformers have thus seen tremendous success, highlighting the gap between theoretical issues and practical successes. Nonetheless, theoretical issues such as those covered in this article are still relevant and significant. These issues undeniably put a ceiling on how much existing models can improve, and must be studied and addressed as both research and industry move to create increasingly more powerful models.

Overall, research into these phenomena are very recent, and therefore sparse. Existing solutions tend to be ad-hoc and motivated by intuition and observation, rather than grounded theory. With GNN practitioners increasingly moving over to transformer research and introducing much needed theoretical clarity into many observed problems of transformers, the field is expected to grow substantially in the coming months and years. 

\bibliographystyle{plainnat}
\bibliography{main}

\end{document}